\documentclass[lettersize,journal]{IEEEtran}

\usepackage[utf8]{inputenc}
\usepackage[T1]{fontenc}
\usepackage{microtype}
\usepackage{textcomp}

\usepackage{amsmath}
\usepackage{amsfonts}
\usepackage{bm}
\usepackage{mathtools}
\usepackage{nicefrac}

\usepackage{graphicx}
\usepackage{subfig} 

\usepackage{array}
\usepackage{tabularx}
\usepackage{booktabs}
\usepackage{multirow}
\usepackage{colortbl}
\usepackage{makecell}

\usepackage{algorithm}
\usepackage{algorithmic}

\usepackage{xcolor}

\definecolor{lightblue}{RGB}{173,216,230}

\definecolor{palepink}{RGB}{255,224,230}

\definecolor{orange}{RGB}{255,165,0}

\usepackage[nocompress]{cite}
\usepackage{hyperref}
\usepackage[capitalise]{cleveref}

\hypersetup{
	colorlinks = true,
	citecolor = blue,
	linkcolor = blue,
	urlcolor = black,
	filecolor = magenta
}

\usepackage{comment}
\usepackage{verbatim}
\usepackage{relsize}

\usepackage{orcidlink}
\usepackage{bbding}
\usepackage{wasysym}

\usepackage{stfloats}
\usepackage{balance}

\def\BibTeX{{\rm B\kern-.05em{\sc i\kern-.025em b}%
\kern-.08em T\kern-.1667em\lower.7ex\hbox{E}\kern-.125emX}}

\begin{document}

\title{MetaRA: Metamorphic Robustness Assessment for Multimodal Large Language Model-based Visual Question Answering Systems}

\author{
Quanxing~Xu\orcidlink{0009-0008-4354-8371},
Yuhao~Tian\orcidlink{0009-0006-7304-4137},
Ling~Zhou\orcidlink{0000-0002-8313-5749},
Xian~Zhong\orcidlink{0000-0002-5242-0467},~\IEEEmembership{Senior~Member,~IEEE},
Xiaohua~Huang\orcidlink{0000-0001-8897-3517},~\IEEEmembership{Senior~Member,~IEEE},
Rubing~Huang\orcidlink{0000-0002-1769-6126},~\IEEEmembership{Senior~Member,~IEEE},
and~Chia-Wen~Lin\orcidlink{0000-0002-9097-2318},~\IEEEmembership{Fellow,~IEEE}
 
\thanks{Manuscript received May 8, 2026. This work was supported in part by the Science and Technology Development Fund of Macau, Macao SAR, under Grants 0069/2025/RIB2 and 0021/2023/RIA1, in part by the National Natural Science Foundation of China under Grant 62271361, in part by the Hubei Provincial Key Research and Development Program under Grant 2024BAB039, and in part by the Hubei Provincial Natural Science Foundation under Grant 2026AFB663. (\textit{Corresponding authors: Ling Zhou and Xian Zhong}.)}

\thanks{Quanxing Xu, Yuhao Tian, and Ling Zhou are with the School of Computer Science and Engineering, Macau University of Science and Technology, Macao SAR 999078, China (e-mail: 3230002299@student.must.edu.mo; 3250010506@student.must.edu.mo; lzhou@must.edu.mo).}

\thanks{Xian Zhong is with the Hubei Key Laboratory of Transportation Internet of Things, School of Computer Science and Artificial Intelligence, Wuhan University of Technology, Wuhan 430070, China, and also with the State Key Laboratory of Maritime Technology and Safety, Wuhan University of Technology, Wuhan 430063, China (e-mail: zhongx@whut.edu.cn).}

\thanks{Xiaohua Huang is with the Oulu School, Nanjing Institute of Technology, Nanjing 210096, China (e-mail: xiaohuahwang@gmail.com).}

\thanks{Rubing Huang is with the School of Computer Science and Engineering, Macau University of Science and Technology, Macao SAR 999078, China, and also with the Macau University of Science and Technology Zhuhai MUST Science and Technology Research Institute, Zhuhai 519099, China (e-mail: rbhuang@must.edu.mo).}

\thanks{Chia-Wen Lin is with the Department of Electrical Engineering, National Tsing Hua University, Hsinchu 30013, Taiwan (e-mail: cwlin@ee.nthu.edu.tw).}
}

\markboth{IEEE TRANSACTIONS ON MULTIMEDIA, 2026}%
{How to Use the IEEEtran \LaTeX Templates}

\maketitle

\begin{abstract}
Visual Question Answering (VQA), as the representative multimodal task, serves as a key benchmark for evaluating the reasoning capabilities of Multimodal Large Language Models (MLLMs). However, existing evaluations largely rely on static datasets and accuracy-based metrics, which fail to capture robustness, consistency, and generalization. Inspired by Metamorphic Testing (MT), we propose \underline{\textbf{Meta}}morphic \underline{\textbf{R}}obustness \underline{\textbf{A}}ssessment (\textbf{MetaRA}), a testing framework that employs Metamorphic Relations (MRs) to systematically probe vulnerabilities in MLLM-based VQA systems. MetaRA generates controlled variations of image-question inputs based on specific MRs and evaluates models across diverse conditions. Applying MetaRA to multiple MLLM-based VQA models across different tasks reveals nuanced failure patterns, including sensitivity to linguistic perturbations, over-reliance on superficial visual cues, and deeper weaknesses in multimodal reasoning. Experimental results demonstrate that MetaRA provides richer diagnostic insights than conventional accuracy metrics, exposing failure modes that remain hidden under standard benchmarks. Overall, this work highlights the need for systematic robustness evaluation in VQA and positions metamorphic assessment as a scalable, model-agnostic approach toward trustworthy multimodal AI. 

\end{abstract}

\begin{IEEEkeywords}
Visual question answering, multimodal large language models, metamorphic testing, metamorphic relations, robustness evaluation
\end{IEEEkeywords}

\maketitle

\section{Introduction}
\label{sec:intro}

\IEEEPARstart{V}{ision}-language tasks serve as representative benchmarks for evaluating models’ capabilities in multimodal learning and visual-linguistic understanding, including visual storytelling~\cite{zhong1}, video captioning~\cite{zhong2,zhong3}, and Visual Question Answering (VQA)~\cite{vqa}. As one of the most fundamental vision-language tasks, VQA requires generating accurate natural-language answers to a question about images and has therefore attracted substantial attention in recent years. This growing interest reflects a broader shift in the image processing community from conventional ``bucketed'' recognition problems toward more complex multimodal reasoning challenges~\cite{vqa,vqav2}.

\begin{figure}[!t]
	\centering
	\includegraphics[width = \linewidth]{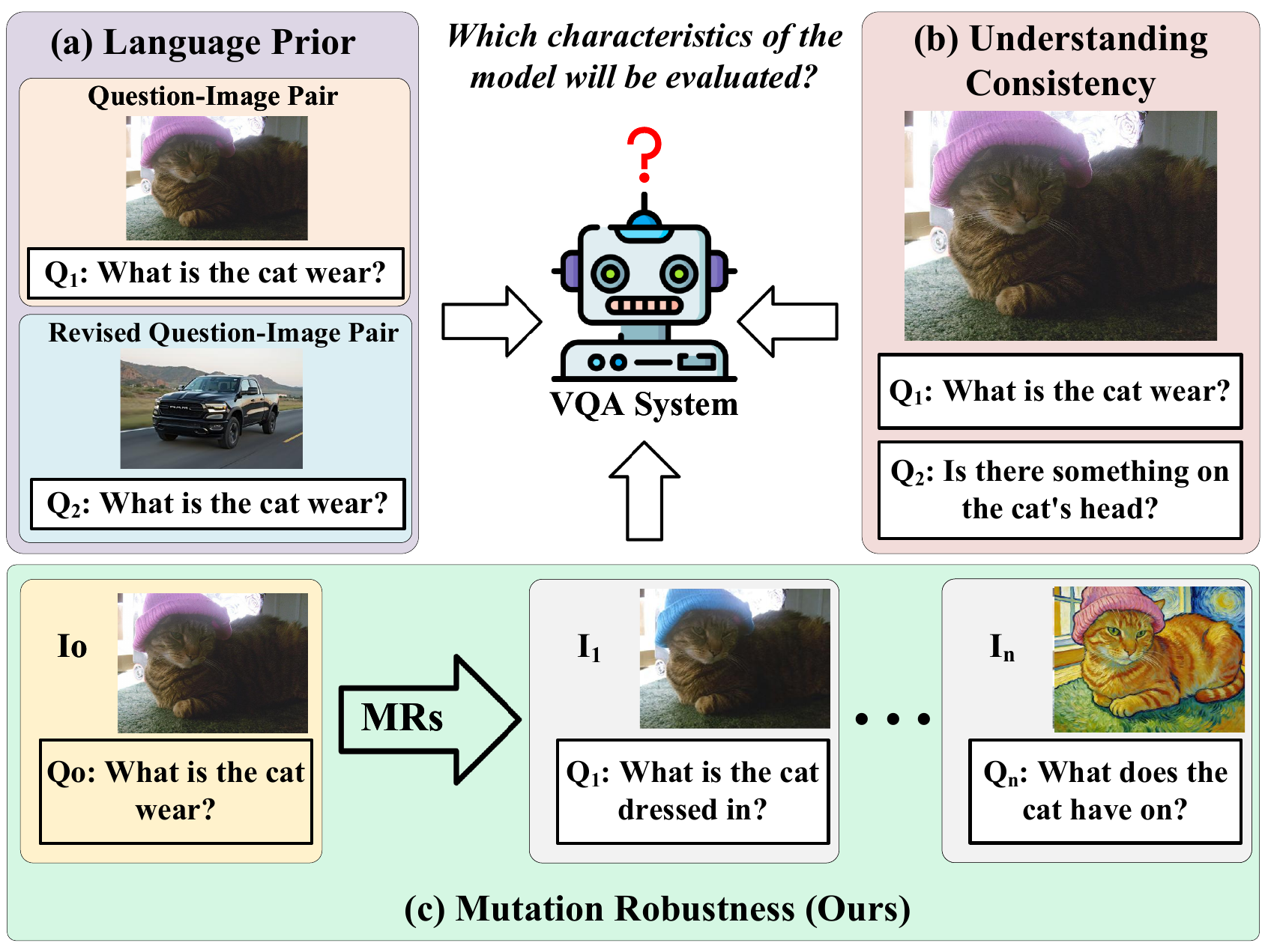}
	 \caption{\textbf{Comparison of robustness evaluation paradigms for VQA.} Existing approaches focus on isolated aspects, including (a) language-bias evaluation and (b) image-understanding evaluation, whereas the proposed method (c) performs a systematic assessment by constructing multiple Metamorphic Relations (MRs) to probe diverse model behaviors.}
	\label{fig1}
\end{figure}


Despite recent progress, robustness remains a critical challenge in VQA~\cite{stability2025}. Although Multimodal Large Language Models (MLLMs) have achieved strong performance in VQA, systematic evaluation of robustness in MLLM-driven VQA systems remains largely underexplored. Given their black-box nature, identifying vulnerabilities in the fusion, understanding, and inference processes is essential for narrowing the interpretability gap in large multimodal models.

Inspired by Metamorphic Testing (MT)~\cite{mt} in software testing, we introduce a new framework to assess the robustness of MLLM-based VQA systems. MT is built upon Metamorphic Relations (MRs), which specify expected logical relationships between inputs and outputs, enabling effective evaluation without explicit ground-truth annotations. Following this principle, we construct a multi-perspective, multi-objective robustness evaluation framework for VQA. As illustrated in \cref{fig1}, our approach moves beyond evaluating isolated model properties to conduct a comprehensive assessment by systematically designing and applying multiple MRs.

Accordingly, we propose \underline{\textbf{Meta}}morphic \underline{\textbf{R}}obustness \underline{\textbf{A}}ssessment (\textbf{MetaRA}), which employs MR-defined test cases to evaluate MLLM-based VQA models and uncover previously unexplored vulnerabilities in handling bimodal inputs. MetaRA focuses on assessing perceptual robustness with respect to both visual-textual interactions and textual information embedded in visual content.

The core contributions of this work are summarized as fourfold:

\begin{itemize}
	\item We propose MetaRA, the first metamorphic testing framework dedicated to robustness assessment of MLLM-based VQA systems, enabling systematic evaluation without relying on ground-truth annotations.
	\item We design a comprehensive suite of MRs tailored for VQA, covering general multimodal comprehension, semantic perception stability, content preservation consistency, and implicit information understanding.
	\item We conduct extensive experiments on mainstream MLLMs across both knowledge-based VQA and OCR-VQA tasks, revealing that state-of-the-art models remain vulnerable to fine-grained visual changes and implicit textual mutations.
	\item We provide in-depth diagnostic insights beyond conventional accuracy metrics, demonstrating that metamorphic assessment serves as a scalable, model-agnostic approach to building trustworthy multimodal AI.
\end{itemize}

Experimental results show that, while current models are relatively robust to global perturbations, they remain vulnerable to fine-grained local visual changes and implicit textual knowledge mutations, revealing weaknesses in multimodal alignment and semantic grounding. Overall, MetaRA provides richer diagnostic insights by systematically uncovering hidden failure modes across diverse VQA tasks, positioning MT-based assessment as a scalable, model-agnostic approach toward trustworthy multimodal AI.

\section{Related Work}
\label{sec:relate}

\subsection{VQA Using MLLMs}

Multimodal Large Language Model (MLLM)-based VQA employs joint vision-language representations to answer natural-language questions grounded in images, thereby improving reasoning and semantic alignment. Existing MLLM-based VQA research mainly focuses on two representative tasks: Knowledge-based VQA (KBVQA)~\cite{kvqa} and OCR-VQA~\cite{ocr-vqa}. KBVQA emphasizes reasoning over visual content combined with external or commonsense knowledge, while OCR-VQA targets the understanding and integration of textual information embedded in images. Despite strong performance on these tasks, robustness evaluation for MLLM-based VQA remains limited. This work addresses this gap by studying robustness testing for MLLM-based VQA under both KBVQA and OCR-VQA settings.

\subsection{Metamorphic Testing for AI}

Metamorphic Testing (MT)~\cite{mt,mtsurvey} is a software testing technique that detects faults by checking whether expected relations hold across multiple inputs and their outputs, thereby alleviating the oracle problem when exact ground truth is unavailable. This property makes MT particularly suitable for MLLMs, where complex cross-modal reasoning can conceal subtle errors. Recent studies have applied MT as a scalable framework for evaluating the reliability of multimodal AI systems~\cite{mtinvqa,reic} across diverse, realistic variations. Building on these efforts, our work employs carefully designed Metamorphic Relations (MRs) to systematically probe vulnerabilities in MLLM-based VQA models.

\subsection{Robustness Evaluation in VQA}

Robustness evaluation has become an indispensable approach to ensuring the reliability and real-world deployment of VQA systems, particularly those built on MLLMs. Conventional VQA evaluation mainly relies on accuracy-based metrics over fixed datasets, which fail to expose model vulnerabilities such as over-reliance on language priors, sensitivity to input perturbations, and unstable multimodal alignment~\cite{stability2025}. Early works assess robustness through unimodal perturbations, including visual noise injection, geometric transformations, and textual paraphrasing, to verify output consistency under mild input changes~\cite{mtinvqa,reic}. 
Nevertheless, these methods typically evaluate robustness from isolated perspectives and lack a unified framework for systematically probing cross-modal defects across diverse transformations. Moreover, most approaches depend on explicit ground-truth answers, limiting their applicability to complex MLLM-based VQA systems with black-box inference mechanisms. In contrast, metamorphic testing offers a promising solution by defining expected relations between inputs and outputs without requiring ground-truth annotations~\cite{mtsurvey}. Building on this paradigm, our MetaRA framework constructs dedicated metamorphic relations to generate controllable visual-textual variations, enabling comprehensive, model-agnostic robustness assessment for MLLM-driven VQA systems.

\begin{figure*}[!t]
	\centering
	\includegraphics[width = \linewidth]{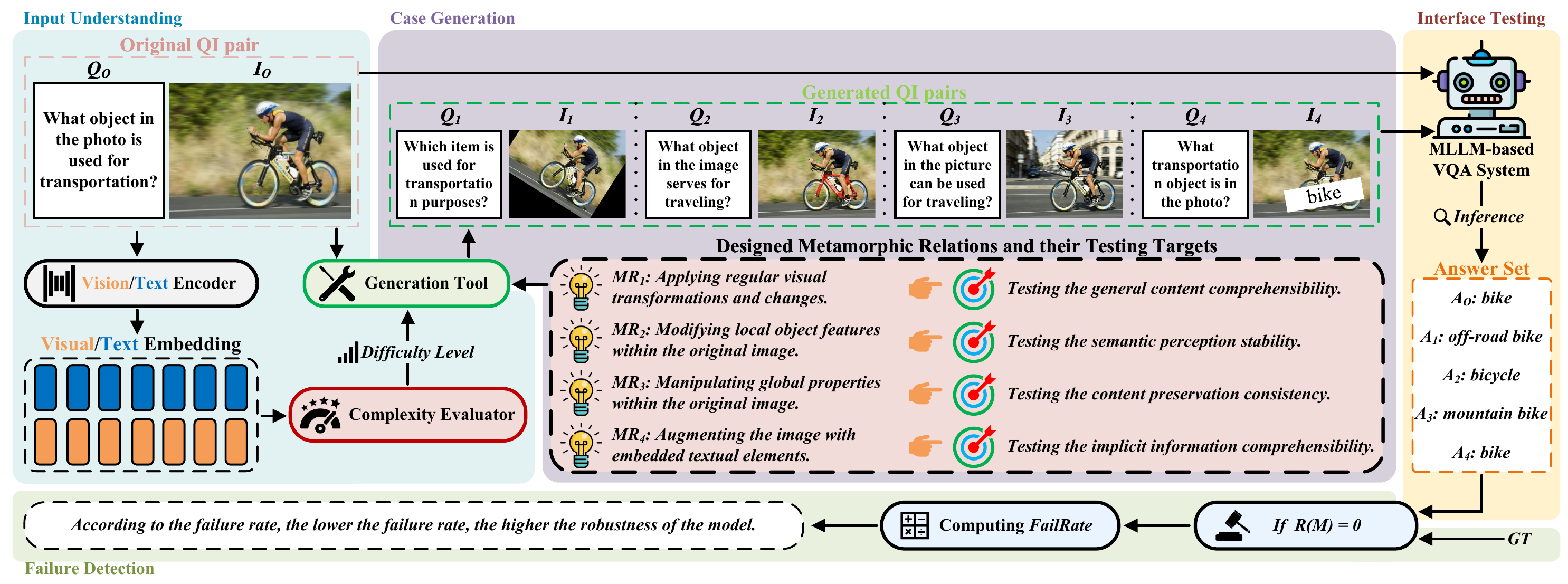}
	\caption{\textbf{Overview of the MetaRA framework.} The framework consists of four stages: Input Understanding, Case Generation, Inference Testing, and Failure Detection, which respectively estimate input difficulty, construct transformation ensembles, evaluate models via systematic testing, and analyze potential defects.}
	\label{fig2}
\end{figure*}

\section{Proposed Method}
\label{sec:metho}

\cref{fig2} summarizes the proposed MetaRA framework. We first introduce the MT-based concept (see \cref{sec:3.1}), then describe the overall procedure (see \cref{sec:3.2}), present MR design (see \cref{sec:3.3}) and case generation tools (see \cref{sec:3.4}), and finally detail failure detection (see \cref{sec:3.5}).

\subsection{Concept}
\label{sec:3.1}

Metamorphic Testing (MT) in multimodal systems provides a practical way to evaluate robustness and reliability when ground-truth outputs are difficult to define across heterogeneous inputs such as text and images. Instead of requiring exact expected answers, it leverages MRs, logical consistency properties that should hold under transformations across modalities. For instance, in VQA systems~\cite{mtinvqa}, rephrasing a question with synonyms or slightly altering an image (\textit{e.g.}, rotation or brightness adjustment) should not change the semantic correctness of the answer; in Image Captioning (IC) systems~\cite{reic}, employing systematic image-level reductions and checking whether generated captions remain semantically consistent can be used to evaluate model robustness.

Given that MLLMs are typically black-box systems that cannot be fully interpreted, especially in multimodal tasks, their performance can be unstable and may even exhibit anomalous behavior. Motivated by prior work on MT in AI, we conduct a case study to comprehensively evaluate the robustness of MLLM-based VQA systems. 

Formally, let $M$ denote an MLLM-based VQA model, and let $x = (I,Q)$ denote a Question-Image (QI) pair consisting of an image $I$ and a question $Q$. Let $x_o = (I_O,Q_O)$ denote the original input, and let $\{x_{t,i}\}_{i = 1}^{n}$ denote the transformed test cases generated by applying MR-guided transformations. Let $C(M,x)$ denote a correctness function that evaluates whether the prediction of model $M$ on input $x$ is consistent with the expected answer (\textit{e.g.}, the ground-truth answer or the MR-consistent output). 

We establish the following criterion: a tested model $M$ shall be recognized as robust if and only if it correctly responds to all given inputs (including the original input and all transformed cases); otherwise, it shall be deemed lacking in robustness. Accordingly, the robustness evaluation function can be formulated as:
\begin{align}
\textit{Robust} \left(M \right) = 
\begin{cases}
1, & \mathrm{if } C \left(M, x_o \right) \land \bigwedge_{i = 1}^{n}C \left(M, x_{t,i} \right), \\
0, & \mathrm{if } \neg C \left(M, x_o \right) \lor \bigvee_{i = 1}^{n}\neg C \left(M, x_{t,i} \right),
\end{cases}
\end{align}
where, $C(M,x_o)$ ensures correctness on the original input, while $\bigwedge_{i = 1}^{n} C(M,x_{t,i})$ enforces consistency across all transformed inputs.

\subsection{MetaRA}
\label{sec:3.2}

We establish an evaluation framework for MLLM-based VQA methods, grounded in the above principle. As shown in \cref{fig2}, our procedure comprises four core processes: Input Understanding, Case Generation, Inference Testing, and Failure Detection, which assess the difficulty of the original input, construct transformation schemes, systematically evaluate models, and analyze potential defects, respectively.


Specifically, when a QI pair $x_o = (I_O,Q_O)$ is fed into the framework, the Complexity Evaluator (CE) first assesses the inference difficulty based on the relevance between the image and the question, as well as the intrinsic content of both modalities. Although existing MLLMs are capable of performing relatively accurate inference on QI pairs with complex content and high answering difficulty (\textit{e.g.}, commonsense reasoning and knowledge-based answering), the difficulty level of QI pairs can serve as a reference metric to reflect the confidence level of the model’s current inference process. In other words, the easier a QI pair is, the higher the probability that the model will yield a correct result; conversely, the more difficult a QI pair is, the lower this probability becomes. Furthermore, this preliminary step is a prerequisite for subsequent procedures, in which the test case generation module constructs cases tailored to the assessed difficulty level.

Upon completion of the difficulty assessment, we generate test cases based on the input QI pair using a vision-text editing tool. For image mutation, inspired by the work of Wang \textit{et al.}~\cite{visualdeeplearningtesting} and Wang \textit{et al.}~\cite{clipinmirror}, we adopt a variety of transformation schemes to evaluate the ability of MLLMs to perceive image semantics. These schemes include horizontal flipping, style transfer, local feature manipulation, modification of scene information, and geometric transformations (\textit{e.g.}, scaling, rotation, and translation). For the question mutation task, given the limited variability in question vocabulary and the relatively low ambiguity of linguistic information, we employ conventional sentence-variation methods, such as synonymous rewriting and constituent replacement.

It is worth noting that image information may convey non-unique semantic meanings, and the complexity of image content tends to increase with the number of applied transformation schemes. Therefore, during the image transformation process, the number and types of transformations are selectively adjusted based on the pre-assessed difficulty level of the QI pair, ensuring that the generated test cases remain comparable in difficulty to the original input. The selection of mutation schemes is determined by the corresponding MRs, as detailed in \cref{sec:3.3}.

Subsequently, a systematic test suite consisting of the original input $x_o$ and the generated test cases $\{x_{t,i}\}_{i = 1}^{n}$ is fed into the model $M$ in a single batch (analogous to sequential examples in a prompt). The model produces predictions $\hat{y}_o = M(x_o)$ and $\hat{y}_{t,i} = M(x_{t,i})$. Considering previous studies~\cite{ordermatter} on the influence of instance input order, we conduct repeated tests with different input sequences for each input set to evaluate the model's sensitivity to ordering effects.

Finally, we perform an in-depth analysis of the model's outputs. By examining incorrect responses, we identify latent flaws and further analyze MLLMs' characteristics in multimodal reasoning and image-text understanding tasks. The detailed procedure is illustrated in Algorithm~\ref{metara}.

\begin{algorithm}[!t]
\caption{MetaRA}
\small
\label{alg:framework}
\begin{algorithmic}[1]
\REQUIRE Model $M$, original input $x_{\mathrm{ori}}$, set of transformation schemes $\mathcal{T}$

\ENSURE Robustness assessment result

\STATE \textit{// Establish baseline and estimate input difficulty}
\STATE \textbf{Input Understanding:} Analyze $x_{\mathrm{ori}}$ to determine semantic complexity and expected ground-truth answer $y_{\mathrm{ori}}$

\STATE \textit{// Generate MR-consistent variations}
\STATE \textbf{Case Generation:} Construct test cases $\{x_{\mathrm{add},i}\}_{i = 1}^{n}$ by applying transformations $t_i \in \mathcal{T}$

\STATE \textit{// Collect outputs under systematic perturbations}
\STATE \textbf{Inference Testing:} Evaluate $M$ on $x_{\mathrm{ori}}$ and $\{x_{\mathrm{add},i}\}$, obtaining predictions $\hat{y}_{\mathrm{ori}}$ and $\{\hat{y}_{\mathrm{add},i}\}$

\STATE \textit{// Check consistency across all cases}
\STATE \textbf{Failure Detection:} 
\IF{$\hat{y}_{\mathrm{ori}} = y_{\mathrm{ori}}$ \AND $\hat{y}_{\mathrm{add},i} = y_{\mathrm{add},i}$ for all $i$}
 \STATE \textit{// All predictions satisfy MR constraints}
 \STATE Mark $M$ as \textbf{Robust}
\ELSE
 \STATE \textit{// Identify violation patterns}
 \STATE Mark $M$ as \textbf{Not Robust Enough} and record failure types
\ENDIF
\end{algorithmic}
\label{metara}
\end{algorithm}

\subsection{Design of MRs}
\label{sec:3.3}
 
At the core of MetaRA, MRs define specific transformation strategies for uncovering potential weaknesses in MLLM-based VQA models. In this work, we construct a set of MRs according to the testing objectives and determine the corresponding transformation operations, thereby completing the generation of test cases.

Specifically, regardless of the difficulty level of the original QI pairs, each input is subjected to four types of MRs, and four corresponding categories of test cases are generated through mutation operations. These four MRs are defined as follows: $MR_1$ (General Multimodal Content Comprehensibility), $MR_2$ (Multimodal Semantic Perception Stability), $MR_3$ (Multimodal Content Preservation Consistency), and $MR_4$ (Multimodal Implicit Information Comprehensibility). Since many existing studies have already explored unimodal perception capabilities, this work does not conduct additional tests on unimodal variations.

Note that the original QI pair includes the ground-truth answer $GT$. Therefore, the robustness of model $M$ is considered valid only when the predictions for all transformed cases $A_T$ and the prediction for the original input $A_O$ are consistent with $GT$:
\begin{align}
\textit{Robust} \left(M \right) = 1 \iff A_O = A_T = GT.
\end{align}
where $A_O = M(I_O,Q_O)$ and $A_T = \{M(I_{t,i},Q_{t,i})\}_{i = 1}^{n}$ denote the predicted answers.

For the original question $Q_O$ and image $I_O$, $MR_1$ applies both textual and visual transformations simultaneously. The question is paraphrased via operations such as synonym substitution or word reordering, and the image is adjusted via benign transformations such as brightness changes, slight rotations, or scaling. This MR evaluates whether the model can maintain semantic stability under minor perturbations:
\begin{align}
M \left(I_O, Q_O \right) = GT \to M \left(t_b \left(I_O \right), p \left(Q_O \right) \right) = GT,
\end{align}
where $t_b(\cdot)$ denotes benign image adjustments and $p(\cdot)$ denotes paraphrasing.

The other three MRs adopt the same textual transformation $p(Q_O)$ but introduce different forms of image mutation to probe distinct aspects of multimodal robustness. $MR_2$ modifies local object features within $I_O$, such as changing color shades or fine-grained textures, to assess sensitivity to detailed visual perception:
\begin{align}
M \left(I_O, Q_O \right) = GT \to M \left(t_\ell \left(I_O \right), p \left(Q_O \right) \right) = GT,
\end{align}
where $t_\ell(\cdot)$ denotes local object-level transformations that preserve object identity and spatial layout.

$MR_3$ manipulates global image properties, such as background information or artistic style (\textit{e.g.}, rendering the image in a Van Gogh style), thereby testing whether the model can preserve global semantic consistency under domain shifts:
\begin{align}
M \left(I_O, Q_O \right) = GT \to M \left(t_g \left(I_O \right), p \left(Q_O \right) \right) = GT,
\end{align}
where $t_g(\cdot)$ denotes global transformations that do not alter scene semantics.

$MR_4$ augments the image with embedded textual elements (\textit{e.g.}, captions or labels) to evaluate whether the model can correctly interpret and align visual-textual relationships:
\begin{align}
M \left(I_O, Q_O \right) = GT \to M \left(\mathrm{overlay} \left(I_O, T \right), p \left(Q_O \right) \right) = GT,
\end{align}
where $\mathrm{overlay}(I_O, T)$ denotes the operation of embedding visually readable text $T$ into the image.

\begin{table}[!t]
	\centering
	\caption{\textbf{Metamorphic relations (MRs) and their corresponding multimodal transformation operations.} Each MR targets a specific aspect of robustness and is instantiated through a combination of question paraphrasing and image-level transformations.}
	\begin{tabularx}{\linewidth}{X|X}
	\toprule[1.1pt]
	{\textbf{MRs (Testing Target)}} & {\textbf{Mutation Operation}} \\ 
	\midrule
	General Content Comprehensibility & \textit{question paraphrasing} + benign visual modifications \\
	\midrule
	Semantic Perception Stability & \textit{question paraphrasing} + local feature modifications \\ 
	\midrule
	Content Preservation Consistency & \textit{question paraphrasing} + background or style transformations \\
	\midrule
	Implicit Information Comprehensibility & \textit{question paraphrasing} + visual-textual entailment augmentation \\
	\bottomrule[1.1pt]
	\end{tabularx}
	\label{tab:mr}
\end{table}

Together, as illustrated in \cref{tab:mr}, these four metamorphic relations provide a structured methodology for evaluating MLLMs in VQA by systematically probing their robustness to paraphrasing, fine-grained perception, global semantic consistency, and cross-modal alignment.

\subsection{Generation Tools}
\label{sec:3.4}


To support the MRs defined in \cref{sec:3.3} and to realize the multimodal transformations described above, we construct a practical mutation pipeline that uses transformer-based models for question mutation and a combination of lightweight Python programs (OpenCV and Pillow) and Stable Diffusion–based generative models for image mutation. Specifically, for a given original QI pair $x_o = (I_O,Q_O)$, the transformation functions $t^Q(\cdot)$ and $t^I(\cdot)$ introduced in \cref{sec:3.2} are instantiated through the following tools.

For question mutation, the paraphrasing function $p(Q_O)$ is implemented using a transformer-based text generation model, which performs synonymous rewriting and constituent replacement while preserving the semantic intent of the original question. This ensures that the generated question $Q'_i = p(Q_O)$ remains logically consistent with the original answer.

For image mutation, the transformation functions $t_b(\cdot)$, $t_\ell(\cdot)$, $t_g(\cdot)$, and $\operatorname{overlay}(\cdot,\cdot)$ corresponding to different MRs are realized as follows:

\subsubsection{$MR_1$}
\textit{Benign Adjustment Operations}. The function $t_b(I_O)$ is implemented using OpenCV and Pillow, including global brightness adjustment, rotation, scaling, and minor geometric transformations. These operations preserve the overall scene semantics while introducing mild visual perturbations.

\subsubsection{$MR_2$}
\textit{Local Feature Modification}. The function $t_\ell(I_O)$ focuses on object-level regions. Region masks are first generated via OpenCV-based segmentation and masking; the masked regions are then modified through fine-grained operations (\textit{e.g.}, slight color shifts, texture perturbations, or selective occlusion) and composited back into the original image using Pillow. This enables targeted manipulation of local features while preserving the global structure.

\subsubsection{$MR_3$} 
\textit{Global Style and Background Transformation}. The function $t_g(I_O)$ is implemented using Stable Diffusion and related neural style transfer frameworks. These methods modify global appearance, including background replacement and artistic style transfer, while maintaining the semantic consistency of objects and their relationships.

\subsubsection{$MR_4$}
\textit{Image-Text Entailment Augmentation}. The function $\operatorname{overlay}(I_O, T)$ embeds textual elements $T$ (\textit{e.g.}, captions, labels, or watermarks) into the image using Pillow’s ImageDraw module, with controlled positioning, font, and color. This ensures that the added textual information is visually coherent and can be consistently interpreted by multimodal models.


By combining these tools, the proposed framework realizes a comprehensive mutation pipeline that covers benign adjustments, local feature modifications, global style transformations, and image-text augmentation. Each transformation is designed to balance controllability, reproducibility, and semantic preservation, ensuring that the generated test cases are both meaningful and suitable for robust evaluation of MLLMs in VQA.

\subsection{Failure Detection}
\label{sec:3.5}

After model inference, if $\textit{Robust}(M) = 0$, we further quantify robustness using the failure rate. Let $\{x_{t,i} = (I'_i,Q'_i)\}_{i = 1}^{n}$ denote the set of transformed QI pairs generated from $x_o$, and let $C(M,x)$ denote the correctness function defined in \cref{sec:3.1}. The failure rate is defined as:
\begin{align}
\textit{FailRate} \left(M \right) = 
\frac{1}{n} \sum_{i = 1}^{n} \mathbf{1} \left[ \neg C \left(M, x_{t,i} \right) \right],
\end{align}
where $\mathbf{1}[\cdot]$ is the indicator function that equals $1$ if the condition holds and $0$ otherwise.

The \textit{FailRate} quantifies the proportion of transformed test cases in which the model violates the expected MR-induced consistency. For each transformed input $x_{t,i}$, the predicate $C(M,x_{t,i})$ evaluates whether the model’s prediction $\hat{y}_{t,i} = M(x_{t,i})$ is consistent with the expected answer (i.e., the ground-truth answer $GT$ or MR-consistent output). A lower \textit{FailRate} indicates stronger robustness and stability across the four metamorphic relations, whereas a higher value reveals sensitivity to different types of perturbations, including textual paraphrasing, benign image transformations, local feature modifications, global style shifts, and image-text entailment augmentation.

In addition to this quantitative metric, we further analyze failure cases by categorizing errors by MR type and transformation operation. This enables a more fine-grained diagnosis of model weaknesses in multimodal alignment, semantic grounding, and cross-modal reasoning.

\begin{table*}[!t]
	\centering
	\caption{\textbf{Failure rate ($\mathrm{FR}$) comparison of MLLM-based KBVQA models on \textsc{E-VQA} and \textsc{InfoSeek}.} $\mathrm{FR}$ is computed for each QI pair and averaged over all samples within a dataset, while the final $\mathrm{FR}$ denotes the average across datasets. Lower values indicate stronger robustness under metamorphic perturbations. Note that lower $\mathrm{FR}$ indicate higher robustness of models,} 
	\setlength\tabcolsep{11pt}
	\begin{tabular}{ll|cc|ccc|c}
	\toprule[1.1pt]
	\multirow{2}[2]{*}{\textbf{Method}} & \multirow{2}[2]{*}{\textbf{LLM/MLLM}} & \multicolumn{2}{c|}{\textsc{\textbf{E-VQA}}} & \multicolumn{3}{c|}{\textsc{\textbf{InfoSeek}}} & \multirow{2}[2]{*}{\textbf{Final FR (\%)}} \\
	\cmidrule(lr){3-4} \cmidrule(lr){5-7}
	& & \textbf{Single-Hop} & \textbf{All} & \textbf{Unseen-Q} & \textbf{Unseen-E} & \textbf{All} & \\
	\midrule
	LLaVA-1.5~\cite{llava1.5} & LLaMA-3.1-8B & 16.0 & 16.9 & 8.3 & 8.9 & 7.8 & 12.5 \\
	LLaVA-1.5~\cite{llava1.5} & Vicuna-7B & 16.3 & 16.9 & 9.6 & 9.4 & 9.5 & 12.5 \\
	Qwen2-VL-7B~\cite{qwen2vl} & - & 16.2 & 17.0 & 15.4 & 16.8 & 16.1 & 11.1 \\
	Qwen2.5-VL-3B~\cite{qwen2.5vl} & - & 17.9 & 19.6 & 20.2 & 21.7 & 20.7 & 10.0 \\
	Qwen2.5-VL-7B~\cite{qwen2.5vl} & - & 21.7 & 20.3 & 22.8 & 24.0 & 23.2 & 9.0 \\ 
	LLaVA-NeXT-7B~\cite{llavanext} & - & 22.1 & 20.8 & 23.2 & 23.4 & 23.1 & 9.0 \\
	LLaVA-NeXT-8B~\cite{llavanext} & - & 22.5 & 21.4 & 23.7 & 24.0 & 23.6 & 9.0 \\
	DPR$_{V+T}$~\cite{dprv+t} & Multi-passage BERT & 29.1 & - & - & - & 12.4 & 11.2 \\
	RORA-VLM~\cite{rora-vlm} & Vicuna-7B & - & 20.3 & 25.1 & 27.3 & 26.0 & 9.2 \\
	EchoSight~\cite{echosight} & Mistral-7B/LLaMA-3-8B & 19.4 & - & - & - & 27.5 & 9.1 \\
	EchoSight~\cite{echosight} & LLaMA-3.1-8B & 26.4 & 24.9 & 18.0 & 19.8 & 18.8 & 9.1 \\
	Wiki-LLaVA~\cite{wikillava} & LLaMA-3.1-8B & 18.3 & 19.6 & 28.6 & 25.7 & 27.1 & 8.3 \\
	Wiki-LLaVA~\cite{wikillava} & Vicuna-7B & 17.7 & 20.3 & 30.1 & 27.8 & 28.9 & 8.2 \\
	ReflectiVA~\cite{reflectiva} & LLaMA-3.1-8B & 35.5 & 35.5 & 28.6 & 28.1 & 28.3 & 9.0 \\
	KU-RAG~\cite{kurag} & GPT-4o & 38.3 & - & - & - & 26.1 & 7.1 \\
	MMKB-RAG~\cite{mmkbrag} & Qwen2-VL-7B & 39.7 & 35.9 & 36.4 & 36.3 & 36.4 & 6.6 \\
	VLM-PRF~\cite{vlmprf} & Qwen2.5-VL-7B & 28.9 & 28.6 & 40.0 & 39.4 & 39.5 & 6.4 \\
	VLM-PRF \textit{w/} RL~\cite{vlmprf} & InternVL3-8B & 40.1 & 39.2 & 43.5 & 42.1 & 42.5 & 5.8 \\
	\bottomrule[1.1pt]
	\end{tabular}
	\label{tab:main1}
\end{table*}

\section{Experimental Results}
\label{sec:experi}

We reorganize the experimental study to provide a clearer and more structured evaluation of MetaRA. We first introduce the evaluation protocol and metrics, followed by datasets and implementation details. We then present quantitative results, MR-level diagnostic analysis, and cross-task comparisons. Finally, we discuss limitations revealed by the experiments.

\subsection{Evaluation Protocol and Metrics} 

MetaRA is designed to conduct a systematic robustness evaluation of MLLM-based VQA models through MR-guided test cases. In addition to standard VQA accuracy~\cite{vqa}, we adopt the failure rate ($\mathrm{FR}$) defined in \cref{sec:3.5} as a primary robustness metric.

Different from conventional evaluation, a lower $\mathrm{FR}$ indicates stronger robustness, while a higher $\mathrm{FR}$ reflects the model’s vulnerability under controlled perturbations. $\mathrm{FR}$ captures failures in multimodal semantic perception, content preservation, and visual-textual alignment that are not reflected by accuracy alone. Therefore, accuracy and $\mathrm{FR}$ are jointly reported to provide complementary insights into both predictive performance and robustness.

\subsection{Datasets and Tasks}

Given the widespread application of MLLMs in both knowledge-intensive and text-centric scenarios, we conduct experiments on two representative VQA task categories: KBVQA and OCR-VQA.

For KBVQA, we adopt \textsc{Encyclopedic-VQA (E-VQA)}~\cite{encyvqa} and \textsc{InfoSeek}~\cite{infoseek}. \textsc{E-VQA} is constructed on a curated Wikipedia-based knowledge base, in which each visual question is associated with supporting textual evidence, enabling knowledge-grounded reasoning. \textsc{InfoSeek} provides a large-scale benchmark for information-seeking VQA, containing both human-annotated and automatically collected samples, with over one million examples for training and evaluation. It emphasizes questions that require external knowledge beyond the visual content, thereby evaluating the model’s ability to integrate visual perception with knowledge reasoning. 

For OCR-VQA, we adopt \textsc{DocVQA}~\cite{docvqa}, \textsc{InfoVQA}~\cite{infovqa}, \textsc{ChartQA}~\cite{chartqa}, and \textsc{TextVQA}~\cite{textvqa}. 
\textsc{DocVQA} contains approximately 50,000 questions over more than 12,000 document images, focusing on structured document understanding and text extraction tasks. 
\textsc{InfoVQA} consists of about 30K questions across 5K infographic images, requiring models to interpret complex visual layouts that combine text, graphics, and numerical information. \textsc{ChartQA} includes over 30K questions on more than 20K charts, covering both human-written and machine-generated queries that require numerical reasoning over visualized data. 
\textsc{TextVQA} contains over 45K questions on around 28K images, emphasizing the ability to read and reason over text embedded in natural scenes. 
These datasets collectively cover diverse text-rich scenarios and form representative benchmarks for evaluating OCR-based multimodal reasoning.

\subsection{Implementation Details}

Following \cref{sec:3.3}, four MRs are employed to generate transformed test cases. For each original QI pair $x_o$, we construct a set $\{x_{t,i}\}_{i = 1}^{n}$ using MR-guided transformations. The number of test cases $n$ is determined by the input difficulty $d(x_o)$ estimated in \cref{sec:3.2}. More difficult inputs are assigned more test cases to ensure sufficient coverage of complex reasoning scenarios.

During inference, we feed the sequence $x_o + x_{t,i}$ ($1 \leq i \leq n$) into the model. To account for order sensitivity~\cite{ordermatter}, we repeat the evaluation under multiple permutations of the input order and report the average $\mathrm{FR}$.

For images containing rich textual content, we follow~\cite{clipinmirror} and replace the local feature modification in $MR_2$ with a mirror flip operation. This setting explicitly evaluates the robustness of models to geometric transformations affecting textual semantics.

\begin{figure}[!t]
	\centering
	\includegraphics[width = \linewidth]{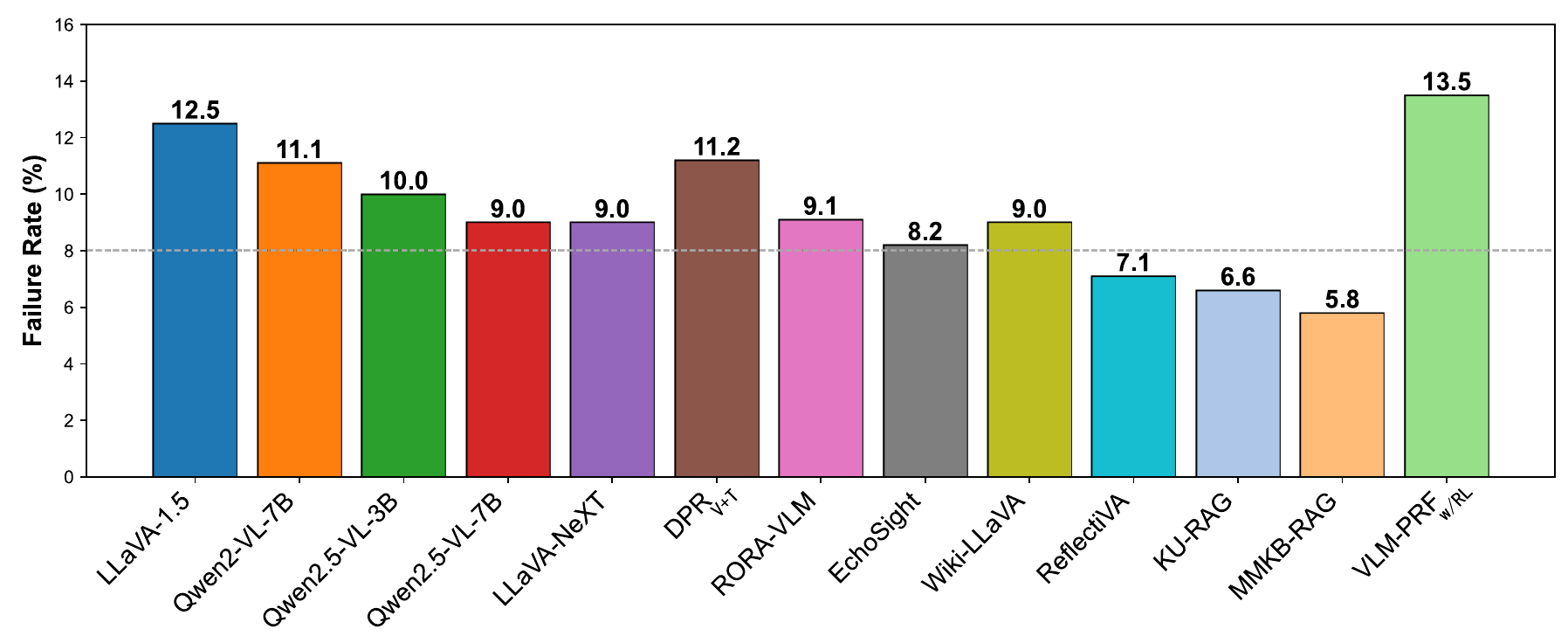}
	\caption{\textbf{Average failure rate ($\mathrm{FR}$) of MLLM-based KBVQA models on \textsc{E-VQA} and \textsc{InfoSeek}.} Lower $\mathrm{FR}$ indicates stronger robustness, and models with better reasoning ability and multimodal alignment generally achieve lower $\mathrm{FR}$.}
	\label{fig3}
\end{figure}

\subsection{Overall Performance Analysis}

\subsubsection{KBVQA Results} 

We evaluate 18 MLLM-based methods on KBVQA tasks. As shown in \cref{tab:main1,fig3}, the overall $\mathrm{FR}$ demonstrates a clear decreasing trend across model evolution. Methods built upon the same backbone exhibit consistent $\mathrm{FR}$ values. Earlier models, such as LLaVA-1.5~\cite{llava1.5} and Qwen2-VL~\cite{qwen2vl}, show relatively high $\mathrm{FR}$, indicating limited robustness. In contrast, more recent models achieve both improved accuracy and reduced $\mathrm{FR}$, suggesting that advances in reasoning capability are generally accompanied by enhanced robustness. VLM-PRF~\cite{vlmprf} achieves the best overall performance, exhibiting the lowest $\mathrm{FR}$.

\begin{table*}[!t]
	\centering
	\caption{\textbf{Failure rate ($\mathrm{FR}$) comparison of MLLM-based OCR-VQA models on \textsc{DocVQA}, \textsc{InfoVQA}, \textsc{ChartQA}, and \textsc{TextVQA}.} $\mathrm{FR}$ is computed for each QI pair and averaged over all samples within each dataset, while the final $\mathrm{FR}$ denotes the average across datasets. Lower values indicate stronger robustness under metamorphic perturbations.}
	\setlength\tabcolsep{16pt}
	\begin{tabular}{lc|c|c|c|c|c}
	\toprule[1.1pt]
	\textbf{Method} & \textbf{Size} & \textsc{\textbf{DocVQA}} & \textsc{\textbf{InfoVQA}} & \textsc{\textbf{ChartQA}} & \textsc{\textbf{TextVQA}} & \textbf{Final FR (\%)} \\
	\midrule
	UReader~\cite{ureader} & 7B & 65.4 & 42.2 & 59.3 & 57.6 & 14.2 \\
	InternVL2~\cite{internvl} & 8B & 91.6 & 74.8 & 83.3 & 77.4 & 9.1 \\
	LLaVA-1.5~\cite{llava1.5} & 7B & 50.3 & 31.6 & 39.4 & 59.0 & 14.4 \\
	DocOwl-1.5~\cite{docowl-1.5} & 8B & 81.6 & 50.4 & 70.5 & 68.8 & 9.1 \\
	DocOwl-1.5-Chat~\cite{docowl-1.5} & 8B & 82.2 & 50.7 & 70.2 & 68.6 & 9.0 \\
	CogAgent~\cite{cogagent} & 17B & 81.6 & 44.5 & 68.4 & 76.1 & 8.2 \\
	Monkey~\cite{monkey} & 9B & 66.5 & 36.1 & 65.1 & 67.6 & 12.1 \\
	TextHawk~\cite{texthawk} & 7B & 76.4 & 50.6 & 66.6 & 59.3 & 11.6 \\
	TextHawk2~\cite{texthawk2} & 7B & 89.6 & 67.8 & 81.4 & 75.1 & 8.0 \\
	HRVDA~\cite{hrvda} & 7B & 72.1 & 43.5 & 67.6 & 73.3 & 9.3 \\
	Vary~\cite{vary} & 7B & 76.3 & 45.0 & 66.1 & 60.6 & 9.2 \\
	MM1.5~\cite{mm1.5} & 7B & 88.1 & 59.5 & 78.6 & 76.8 & 10.3 \\
	Marten~\cite{marten} & 8B & 92.0 & 75.2 & 81.7 & 74.4 & 8.6 \\
	\bottomrule[1.1pt]
	\end{tabular}
	\label{tab:main2}
\end{table*}

\begin{figure}[!t]
	\centering
	\includegraphics[width = \linewidth]{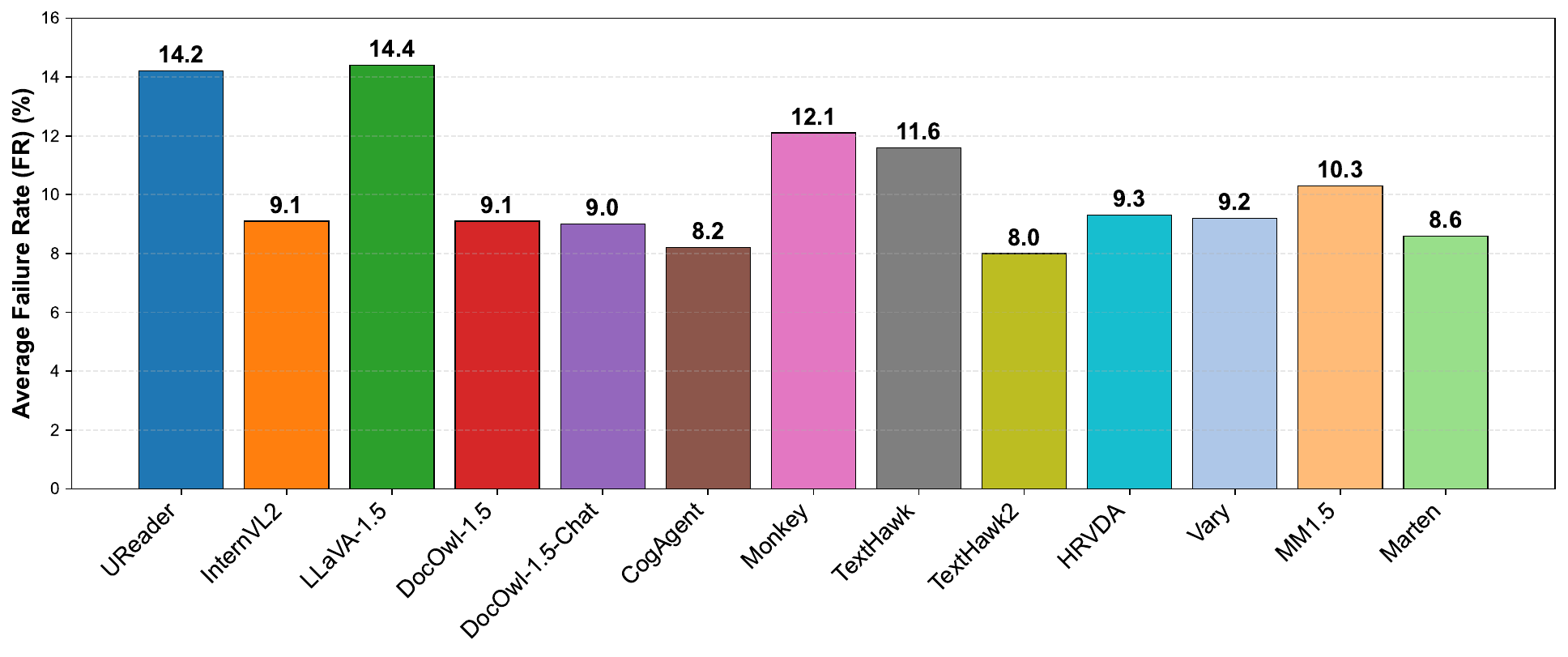}
	\caption{\textbf{Failure rate ($\mathrm{FR}$) comparison of MLLM-based OCR-VQA models across four benchmarks.} Results are reported on \textsc{DocVQA}, \textsc{InfoVQA}, \textsc{ChartQA}, and \textsc{TextVQA}, with the final $\mathrm{FR}$ representing the average across all datasets, where lower values indicate higher robustness.}
	\label{fig4}
\end{figure}

\subsubsection{OCR-VQA Results} 

We further evaluate 13 methods on OCR-VQA tasks. As illustrated in \cref{tab:main2,fig4}, $\mathrm{FR}$ exhibits a fluctuating pattern rather than a monotonic decrease. Even recent models, such as MM1.5~\cite{mm1.5}, maintain relatively high $\mathrm{FR}$. This behavior is mainly attributed to the increased difficulty of OCR-VQA, in which the mirror-flip operation in $MR_2$ significantly disrupts text perception. These results indicate that improvements in robustness in general VQA do not directly transfer to text-intensive scenarios.

\subsection{Consistency Between MetaRA and MetaVQA}

To verify the correctness and effectiveness of the proposed MetaRA, we conduct a consistency experiment comparing our method with the previous MT-based VQA approach proposed in~\cite{mtinvqa} (MetaVQA). We select five mainstream MLLMs and evaluate them on \textsc{E-VQA} (KBVQA) and \textsc{TextVQA} (OCR-VQA) using both methods.

\begin{table*}[!t]
	\centering
	\caption{\textbf{Robustness ranking consistency between the proposed MetaRA and MetaVQA~\cite{mtinvqa}.} A lower $\mathrm{FR}$ indicates stronger robustness, and Rank~1 denotes the most robust model.}
	\setlength\tabcolsep{8pt}
	\begin{tabular}{lc|cc|cc|ccc}
	\toprule[1.1pt]
	\multirow{2}[2]{*}{\textbf{Model}} & \multirow{2}[2]{*}{\textbf{Size}} & \multicolumn{2}{c|}{\textsc{\textbf{E-VQA}}} & \multicolumn{2}{c|}{\textsc{\textbf{TextVQA}}} & \multirow{2}[2]{*}{\textbf{MetaRA Rank}} & \multirow{2}[2]{*}{\textbf{MetaVQA Rank}} & \multirow{2}[2]{*}{\textbf{Rank Consistency}} \\
	\cmidrule(lr){3-4} \cmidrule(lr){5-6}
	& & \textbf{MetaVQA} & \textbf{MetaRA} & \textbf{MetaVQA} & \textbf{MetaRA} & & & \\
	\midrule
	LLaVA-1.5~\cite{llava1.5} & 7B & 28.7 & 16.0 & 37.2 & 59.0 & 5 & 5 & \checkmark \\
	Qwen2-VL~\cite{qwen2vl} & 7B & 27.1 & 16.2 & 35.8 & 60.6 & 4 & 4 & \checkmark \\
	LLaVA-NeXT~\cite{llavanext} & 7B & 22.3 & 22.1 & 31.5 & 73.3 & 3 & 3 & \checkmark \\
	Qwen2.5-VL~\cite{qwen2.5vl} & 7B & 21.6 & 21.7 & 30.1 & 67.6 & 2 & 2 & \checkmark \\
	InternVL2~\cite{internvl} & 8B & -- & -- & 26.3 & 77.4 & -- & -- & -- \\
	\bottomrule[1.1pt]
	\end{tabular}
	\label{tab:mtconsistency}
\end{table*}

\begin{figure}[!t]
	\centering
	\includegraphics[width = \linewidth]{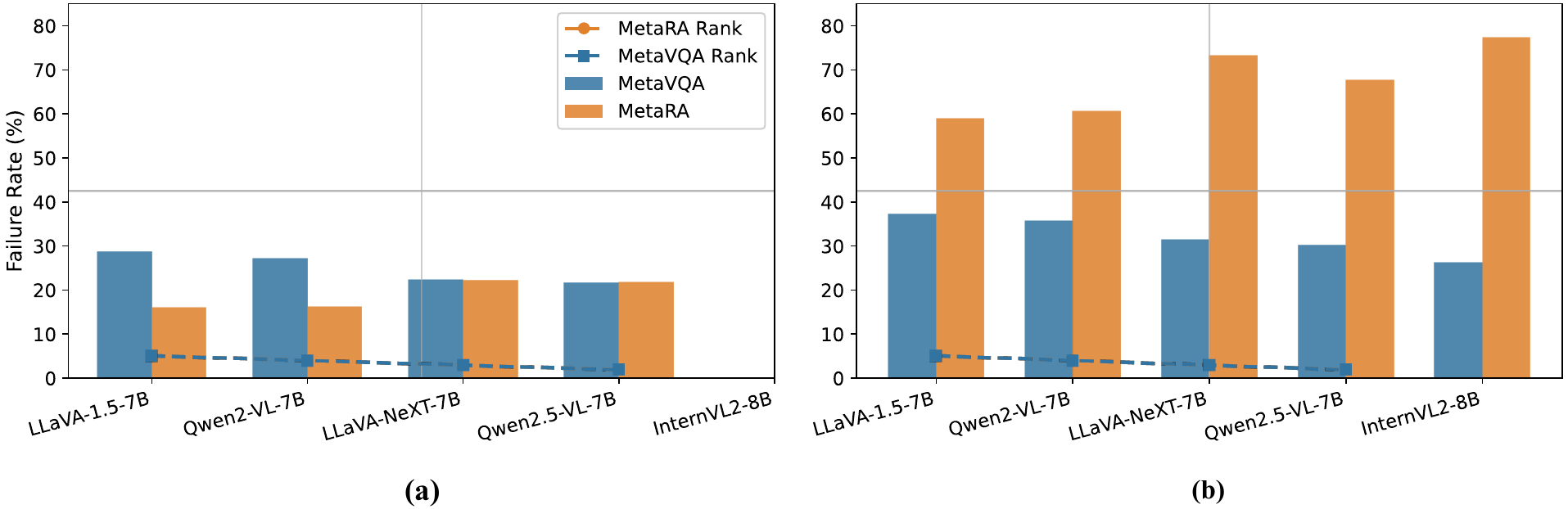}
	\caption{\textbf{$\mathrm{FR}$ comparison and ranking consistency between MetaVQA (a) and MetaRA (b) on \textsc{E‑VQA} and \textsc{TextVQA}.} Note that lower $\mathrm{FR}$ indicates stronger robustness. The dashed lines show the perfectly consistent ranking trend.}
	\label{fig5}
\end{figure}

\cref{tab:mtconsistency} shows the ranking consistency between the proposed MetaRA and MetaVQA~\cite{mtinvqa}.
Specifically, \textit{MetaRA Rank} refers to the robustness rank of models sorted by $\mathrm{FR}$ values obtained from our MetaRA framework,
and \textit{MetaVQA Rank} indicates the rank obtained from the baseline MT method MetaVQA.
In both ranking schemes, Rank~1 denotes the most robust model, while a larger rank number indicates weaker robustness.
The $\checkmark$ in the \textit{Rank Consistency} column means that the model achieves exactly the same robustness rank under both MetaRA and MetaVQA evaluation paradigms.
As shown in~\cref{fig5}, all compared models yield fully consistent robustness rankings across MetaRA and MetaVQA on both the \textsc{E-VQA} and \textsc{TextVQA} datasets.
The identical trends of the two methods demonstrate that MetaRA not only inherits the core logic of MT but also provides a more comprehensive and fine-grained robustness diagnosis, confirming its correctness and reliability.

\begin{table*}[!t]
	\centering
	\caption{\textbf{Failure rate ($\mathrm{FR}$) of LLaVA-1.5 (7B) across KBVQA and OCR-VQA benchmarks.} Results are reported on six datasets, illustrating the model’s heterogeneous robustness across task types, with higher $\mathrm{FR}$ observed on text-intensive OCR-VQA benchmarks.}
	\setlength\tabcolsep{17.5pt}
	\begin{tabular}{l|cc|cccc} 
	\toprule[1.1pt]
	\multirow{2}[2]{*}{\textbf{Method}} & \multicolumn{2}{c|}{\textbf{KBVQA}} & \multicolumn{4}{c}{\textbf{OCR-VQA}} \\ 
	\cmidrule(lr){2-3} \cmidrule(lr){4-7}
	& \textsc{\textbf{E-VQA}} & \textsc{\textbf{InfoSeek}} & \textsc{\textbf{DocVQA}} & \textsc{\textbf{InfoVQA}} & \textsc{\textbf{ChartQA}} & \textsc{\textbf{TextVQA}} \\
	\midrule
	LLaVA-1.5~\cite{llava1.5} & 11.4 & 13.6 & 14.2 & 16.8 & 16.6 & 10.0 \\
	\bottomrule[1.1pt]
	\end{tabular}
	\label{tab:llava}
\end{table*}

\begin{figure}[!t]
	\centering
	\includegraphics[width = 0.8\linewidth]{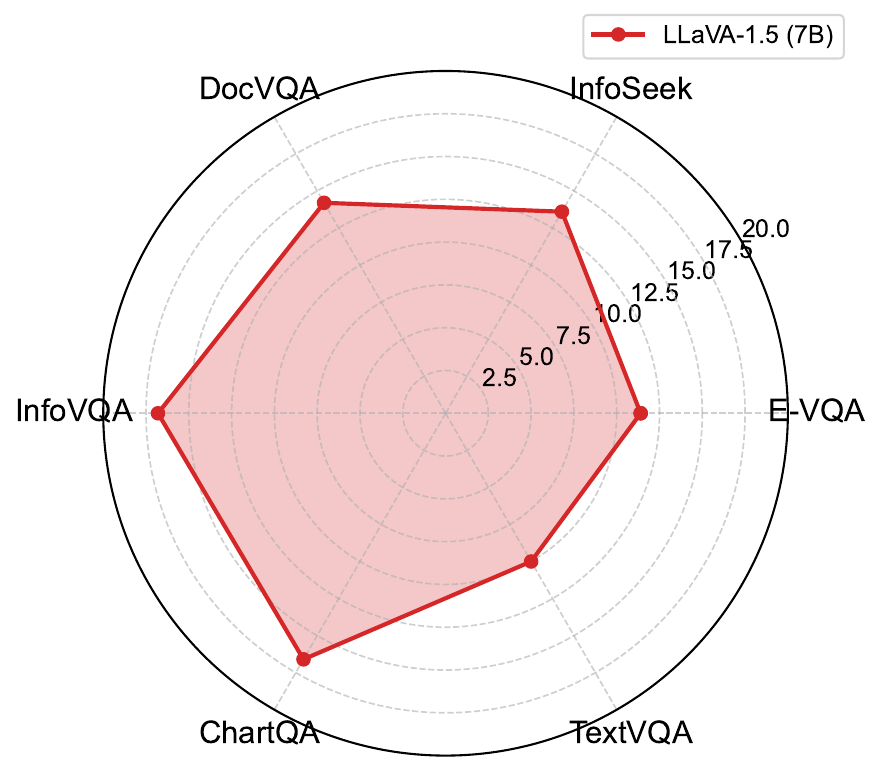}
	\caption{\textbf{Dataset-wise $\mathrm{FR}$ distribution of LLaVA-1.5 (7B) across KBVQA and OCR-VQA benchmarks.} The radar chart highlights higher $\mathrm{FR}$ on \textsc{InfoVQA} and \textsc{ChartQA}, indicating these datasets pose greater robustness challenges.}
	\label{fig6}
\end{figure}

\begin{figure}[!t]
	\centering
	\includegraphics[width = 0.8\linewidth]{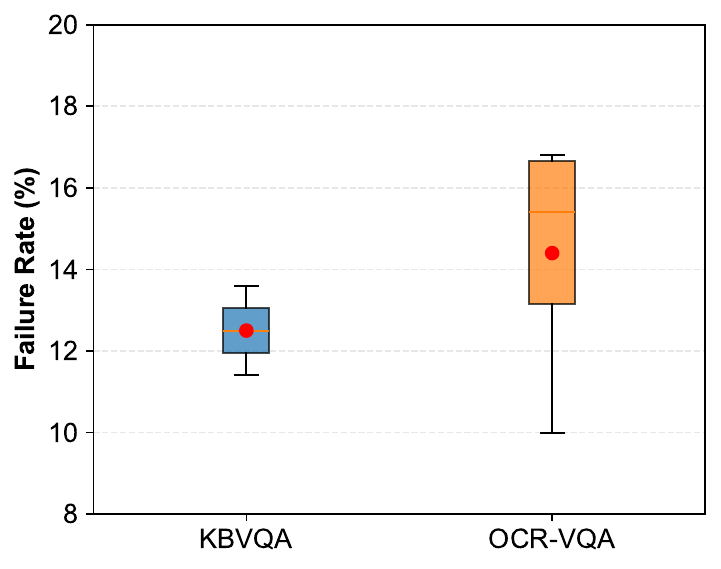}
	\caption{\textbf{$\mathrm{FR}$ distribution comparison between KBVQA and OCR-VQA tasks.} OCR-VQA exhibits a wider distribution range and higher overall $\mathrm{FR}$, indicating increased difficulty in text-intensive multimodal reasoning.}
	\label{fig7}
\end{figure}

\subsection{Cross-Task Robustness Comparison}

To further analyze robustness across task types, we evaluate the same model on both KBVQA and OCR-VQA benchmarks. As shown in \cref{tab:llava,fig6,fig7}, LLaVA-1.5~\cite{llava1.5} exhibits highly heterogeneous $\mathrm{FR}$ distributions across datasets.

The radar chart (see \cref{fig6}) provides a unified view, highlighting \textsc{InfoVQA} and \textsc{ChartQA} as the most challenging datasets with the highest $\mathrm{FR}$. The boxplot (see \cref{fig7}) further shows that OCR-VQA tasks have both a higher average $\mathrm{FR}$ and greater variance than KBVQA. These observations indicate that text-rich reasoning introduces greater instability and sensitivity to perturbations.

\begin{table}[!t]
	\centering
	\caption{\textbf{Average $\mathrm{FR}$ across four MRs for KBVQA and OCR-VQA.} The reported values are obtained by averaging $\mathrm{FR}$ of each MR over all corresponding datasets, highlighting differences in robustness across testing targets.}
	\setlength\tabcolsep{9pt}
	\begin{tabular}{l|cc}
	\toprule[1.1pt]
	\multirow{2}[2]{*}{\textbf{MRs (Testing Target)}} & \multicolumn{2}{c}{\textbf{Average FR (\%)}} \\ 
	\cmidrule(lr){2-3}
	& \textbf{KBVQA} & \textbf{OCR-VQA} \\
	\midrule
	General Content Comprehensibility & 6.4 & 7.1 \\
	Semantic Perception Stability & 7.2 & 11.3 \\
	Content Preservation Consistency & 9.1 & 9.3 \\
	Implicit Information Comprehensibility & 5.8 & 13.5 \\
	\bottomrule[1.1pt]
	\end{tabular}
	\label{tab:avefr}
\end{table}

\subsection{MR-Level Diagnostic Analysis}

\begin{figure}[!t]
	\centering
	\includegraphics[width = \linewidth]{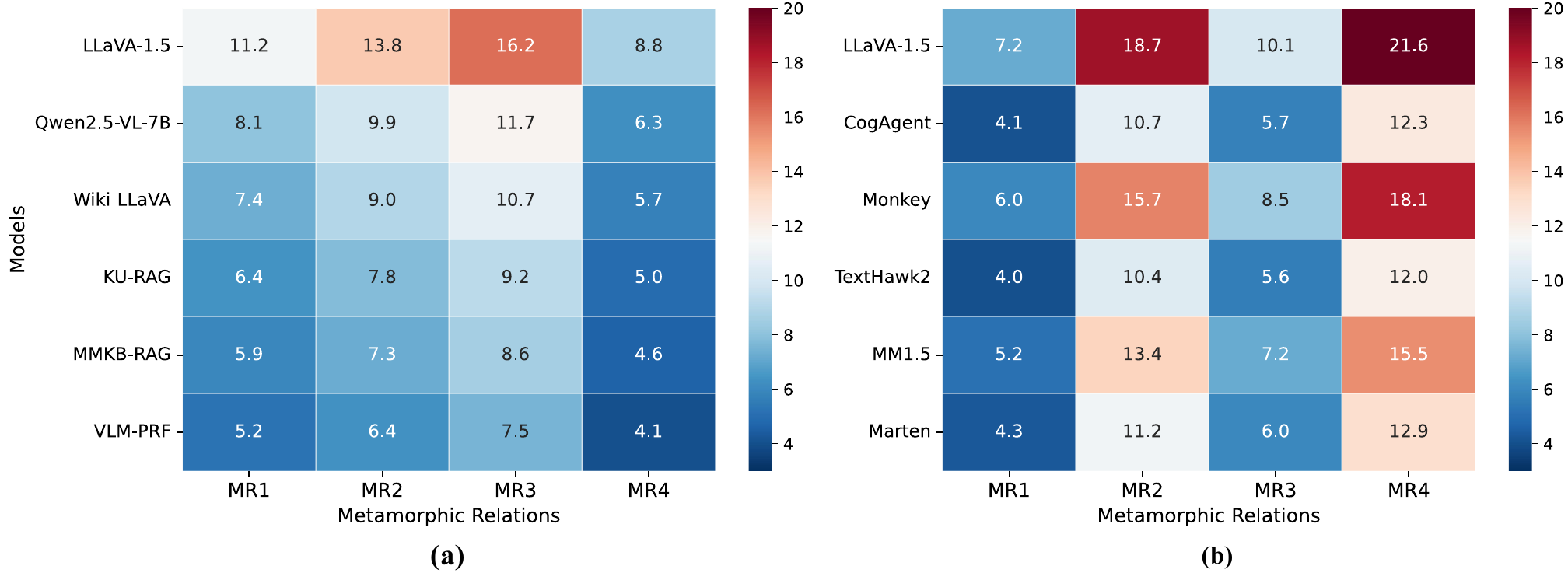}
	\caption{\textbf{MR-specific robustness of models on KBVQA and OCR-VQA tasks.} Note that a darker color indicates a higher $\mathrm{FR}$, implying worse robustness.}
	\label{fig8}
\end{figure}

We further analyze robustness with respect to the four MRs, as shown in \cref{tab:avefr,fig8}. The average $\mathrm{FR}$ indicates that models are relatively robust to benign transformations ($MR_1$) and global style changes ($MR_3$), while exhibiting significantly higher $\mathrm{FR}$ under local perturbations ($MR_2$) and implicit visual-textual reasoning ($MR_4$). Notably, the maximum $\mathrm{FR}$ on OCR-VQA (13.5\%) is nearly 50\% higher than the highest FR on KBVQA (9.1\%), demonstrating that models’ robustness varies significantly with task type, with OCR-VQA presenting a more challenging robustness evaluation scenario.

The models’ robustness across both tasks and all four MRs is further analyzed, with results visualized in a heatmap. This enables an intuitive comparison of failure rates across different settings. As illustrated in \cref{fig8}, vulnerability patterns differ across task types, with a darker color indicating a higher $\mathrm{FR}$, implying worse robustness. While $MR_1$ and $MR_3$ maintain relatively low $\mathrm{FR}$ across both KBVQA and OCR-VQA, $MR_2$ and $MR_4$ lead to substantially higher failure rates, particularly in OCR-VQA. This indicates that local structural changes and implicit visual--textual reasoning introduce more severe robustness degradation in text-centric scenarios.

\begin{figure}[!t]
	\centering
	\includegraphics[width = \linewidth]{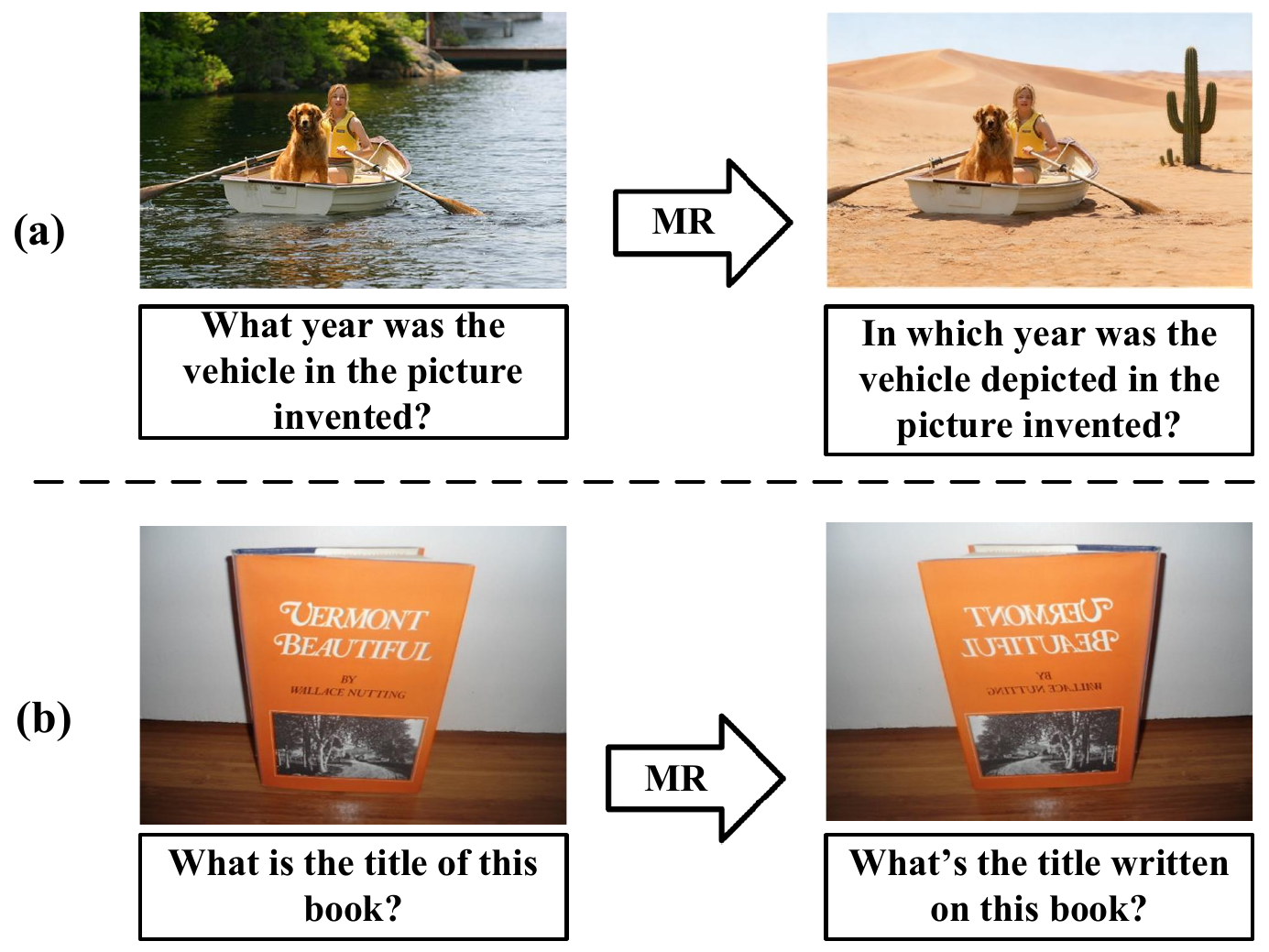}
	\caption{\textbf{Examples of metamorphic mutations in KBVQA and OCR-VQA QI pairs.} The comparison illustrates how different mutation strategies affect visual-textual consistency across the two task types.}
	\label{fig9}
\end{figure}

We attribute this phenomenon to the distinct roles of textual information in images across the two tasks. In OCR-VQA, textual content is explicit and often central to answering questions, making models highly sensitive to geometric distortions and semantic perturbations. This effect is further illustrated in \cref{fig9}. In case (a), although a large portion of the image content is altered, the textual information remains interpretable, and the model is able to produce a correct answer. In contrast, in case (b), the overall scene appears largely unchanged, yet subtle distortions in the embedded textual content lead to incorrect predictions. This observation highlights that OCR-VQA tasks impose stringent geometric and structural constraints on textual visual information, where even minor perturbations can disrupt semantic understanding.

In contrast, KBVQA relies more heavily on external knowledge, with images primarily serving as contextual anchors for knowledge retrieval rather than the main source of semantic content. Consequently, perturbations that affect textual salience or visual precision have a disproportionately larger impact on OCR-VQA than on KBVQA. These findings further confirm that equivalent metamorphic perturbations can lead to task-specific differences in robustness, emphasizing the need for task-aware evaluation strategies.

\subsection{Discussion and Limitation Analysis}

The experimental results reveal several limitations of current MLLM-based VQA models. Although models demonstrate robustness to global perturbations such as background shifts and stylistic transformations, their performance degrades under fine-grained local modifications and implicit textual mutations. This suggests that multimodal alignment and semantic grounding remain fragile in challenging scenarios.

Furthermore, cross-task comparisons indicate that KBVQA models struggle to consistently integrate external knowledge, whereas OCR-VQA models are particularly sensitive to structural and geometric variations in text.

Overall, these findings suggest that, despite strong performance on standard benchmarks, current MLLMs lack robust mechanisms for handling fine-grained visual changes and implicit multimodal semantics. Future work should focus on improving multimodal alignment, enhancing retrieval-augmented reasoning, and developing training strategies that explicitly target robustness under metamorphic transformations.

\section{Conclusion}

We present \underline{\textbf{Meta}}morphic \underline{\textbf{R}}obustness \underline{\textbf{A}}ssessment (\textbf{MetaRA}), a robustness testing framework for MLLM-based VQA that leverages specifically designed MRs to systematically probe model vulnerabilities. By generating controlled variations of QI pairs, MetaRA reveals subtle weaknesses in current methods, including sensitivity to linguistic perturbations, over-reliance on superficial visual cues, and deeper inconsistencies in multimodal association. Experimental results demonstrate that MetaRA provides diagnostic insights beyond conventional accuracy metrics, effectively uncovering failure modes that remain hidden under standard benchmarks. These findings highlight the necessity of robustness-oriented evaluation in multimodal AI. Overall, MetaRA establishes metamorphic testing as a scalable, model-agnostic methodology that facilitates the development of more reliable, interpretable, and trustworthy VQA systems.
Future work will explore more diverse MR design strategies to further improve testing coverage, and investigate evaluation metrics that more precisely quantify the effectiveness and accuracy of metamorphic testing.






\balance
\bibliographystyle{ieeetr}

\bibliography{MetaRA}

\twocolumn[
\begin{@twocolumnfalse}
	\section*{\centering{Supplementary Materials for \\ \emph{MetaRA: Metamorphic Robustness Assessment for Multimodal Large Language Model-based Visual Question Answering Systems\\[30pt]}}}
\end{@twocolumnfalse}
]

\setcounter{page}{1} 
\pagenumbering{arabic} 
\addcontentsline{toc}{section}{Appendices}  

This is the supplementary material accompanying our main paper, ``MetaRA: Metamorphic Robustness Assessment for Multimodal Large Language Model-based Visual Question Answering Systems''. In this supplementary material, we provide additional information that could not be included in the main paper due to space constraints. We further elaborate on the different influences of four MRs in two subsections: Section A presents an intuitive average $\mathrm{FR}$ comparison of four MRs across two VQA tasks, and Section B gives an obvious difference in MR robustness across two VQA tasks. 

\vspace{3mm}

\subsection{Average $\mathrm{FR}$ Comparison}

\begin{figure}[!b]
  \renewcommand{\thefigure}{A}
  \centering
  \includegraphics[width=1\linewidth]{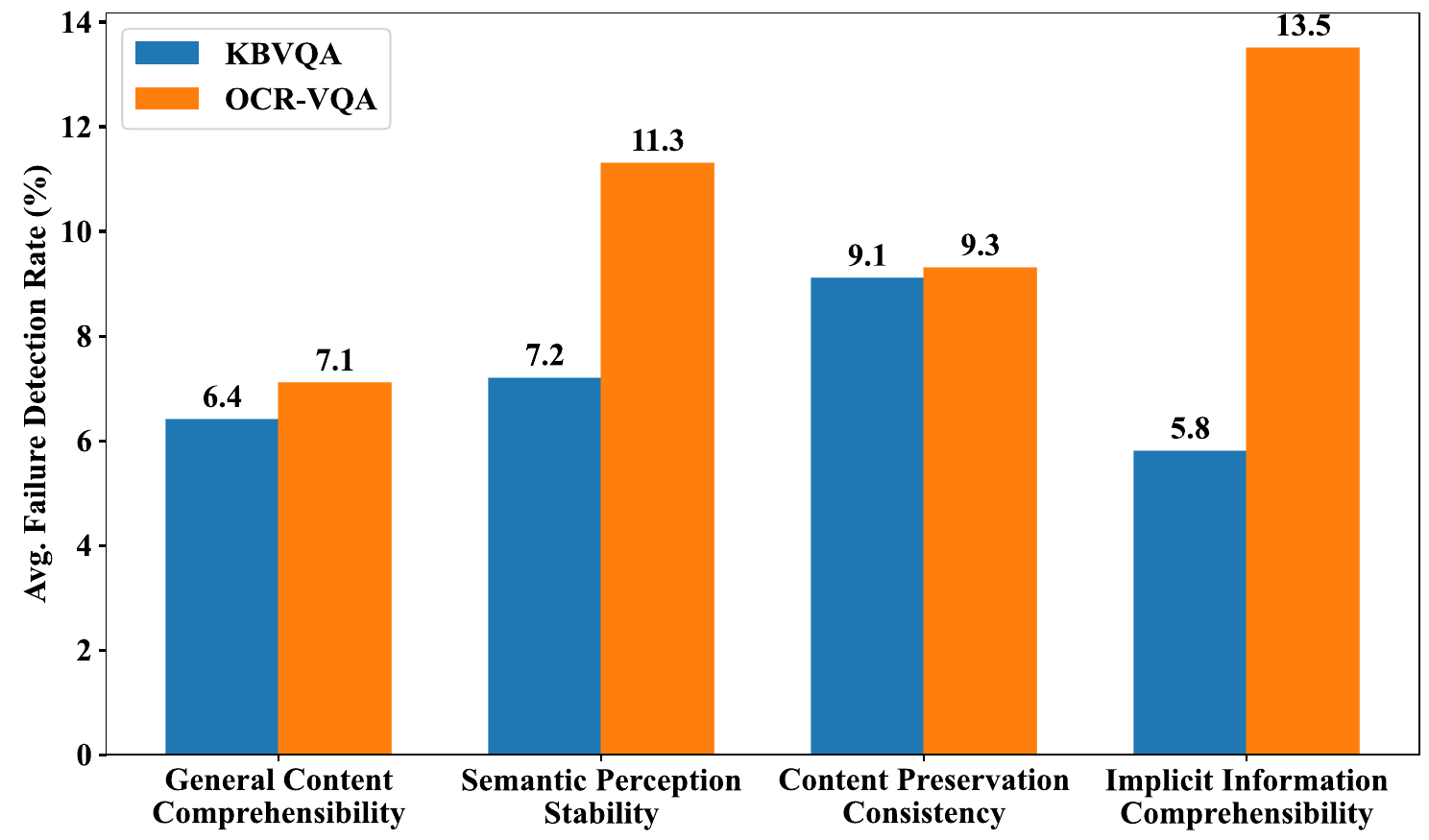}
  \caption{\textbf{$\mathrm{FR}$ comparison across four MRs for KBVQA and OCR-VQA.} Significant differences are observed in semantic perception stability and implicit information comprehensibility, reflecting the varying importance of textual information in the two tasks.
  }
  \vspace{-1mm}
  \label{figsup1}
\end{figure}

\cref{figsup1} compares the average $\mathrm{FR}$ across four MRs for KBVQA and OCR-VQA tasks. The results show that $\mathrm{FR}$ generally increases with the complexity of the robustness aspect being tested, from basic content comprehensibility to implicit information reasoning. Across all MRs, OCR-VQA exhibits consistently higher $\mathrm{FR}$ values than KBVQA, particularly under Semantic Perception Stability (11.3\% vs. 7.2\%) and Implicit Information Comprehensibility (13.5\% vs. 5.8\%).

These significant gaps reflect the differing importance of textual information in the two tasks. KBVQA relies more on general visual and common-sense knowledge, resulting in relatively stable performance even under perturbations. In contrast, OCR-VQA depends heavily on accurate text parsing and contextual reasoning, making it far more vulnerable to subtle changes in semantic and implicit information, as captured by $MR_2$ and $MR_4$.

\subsection{Robustness Comparison}

\begin{figure}[!b]
  \renewcommand{\thefigure}{B}
  \centering
  \includegraphics[width=1\linewidth]{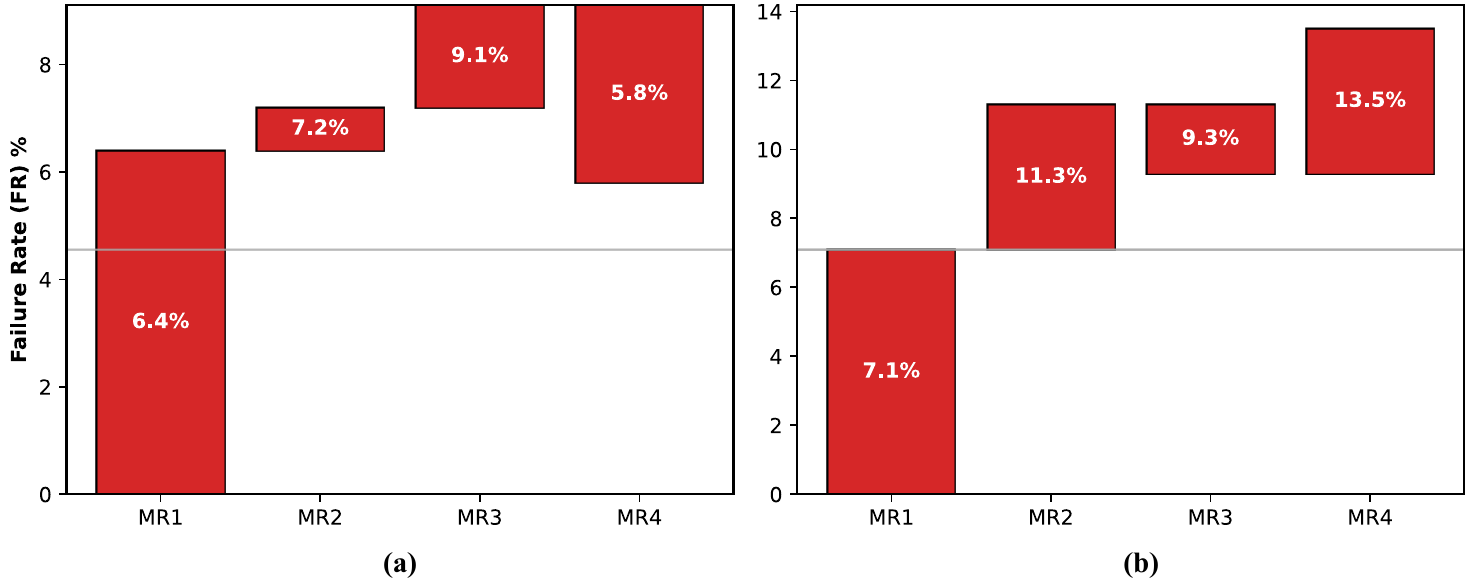}
  \caption{\textbf{MR robustness waterfall plots on KBVQA (a) and OCR-VQA (b).} Each bar represents the $\mathrm{FR}$ under four MRs. The plots highlight the progressive, heterogeneous degradation of robustness across different MRs and tasks.
  }
  \vspace{-1mm}
  \label{figsup2}
\end{figure}

\cref{figsup2} presents waterfall plots of $\mathrm{FR}$ across the four MRs for KBVQA and OCR-VQA, illustrating the progressive degradation of robustness. For KBVQA, $\mathrm{FR}$ rises gradually from 6.4\% ($MR_1$) to 9.1\% ($MR_3$) before dropping to 5.8\% ($MR_4$), indicating moderate and relatively stable fragility across most MRs, with the exception of implicit information perturbations. In contrast, OCR-VQA shows a steeper, more volatile trend, starting at 7.1\% ($MR_1$) and peaking at 13.5\% ($MR_4$), with sharp increases in $MR_2$ and $MR_4$.

This heterogeneous pattern reveals that OCR-VQA models have multiple critical failure points, particularly in semantic perception and implicit text comprehension. The results confirm that robustness degradation is task-dependent and uneven, highlighting the need for multidimensional MR testing to expose the full spectrum of vulnerabilities in VQA models.

\end{document}